\documentclass[letterpaper]{article} % DO NOT CHANGE THIS
\usepackage{aaai25}  % DO NOT CHANGE THIS
\usepackage{times}  % DO NOT CHANGE THIS
\usepackage{helvet}  % DO NOT CHANGE THIS
\usepackage{courier}  % DO NOT CHANGE THIS
\usepackage[hyphens]{url}  % DO NOT CHANGE THIS
\usepackage{graphicx} % DO NOT CHANGE THIS
\usepackage{natbib}  % DO NOT CHANGE THIS AND DO NOT ADD ANY OPTIONS TO IT
\usepackage{caption} % DO NOT CHANGE THIS AND DO NOT ADD ANY OPTIONS TO IT
\usepackage{algorithm}
\usepackage{algorithmic}
\usepackage{booktabs}
\usepackage{multirow}
\usepackage{subcaption}
\usepackage{xcolor}
\usepackage{graphicx}
\usepackage{latexsym}
\usepackage{amsmath}
\usepackage{makecell}
\usepackage{fdsymbol}
\usepackage{pifont}
\usepackage{array}
\usepackage{fancyvrb}
\usepackage{tcolorbox}
\usepackage{tabularx,longtable}
\definecolor{promptbg}{RGB}{230, 230, 250}
\definecolor{responsebg}{RGB}{240, 255, 240}

\newtcolorbox{promptbox}{
  colback=promptbg,
  colframe=blue,
  boxrule=0.5mm,
  sharp corners,
  left=1mm,
  right=1mm,
  top=1mm,
  bottom=1mm,
}

\newtcolorbox{responsebox}{
  colback=responsebg,
  colframe=green,
  boxrule=0.5mm,
  sharp corners,
  left=1mm,
  right=1mm,
  top=1mm,
  bottom=1mm,
}

\definecolor{instructionbg}{RGB}{220, 220, 220}

\newtcolorbox{instructionbox}{
  colback=instructionbg,
  colframe=black,
  boxrule=0.5mm,
  sharp corners,
  left=1mm,
  right=1mm,
  top=1mm,
  bottom=1mm,
}

\usepackage{newfloat}
\usepackage{listings}
\DeclareCaptionStyle{ruled}{labelfont=normalfont,labelsep=colon,strut=off} % DO NOT CHANGE THIS
\floatstyle{ruled}
\newfloat{listing}{tb}{lst}{}
\floatname{listing}{Listing}
\title{Aligning Language Models Using Follow-up Likelihood as Reward Signal}
\author {
    Chen Zhang\textsuperscript{\rm 1}
    \quad Dading Chong\textsuperscript{\rm 2}
    \quad Feng Jiang\textsuperscript{\rm 3,5,6}\thanks{Corresponding Author}  \\
    Chengguang Tang\textsuperscript{\rm 4}
    \quad Anningzhe Gao\textsuperscript{\rm 5}
    \quad Guohua Tang\textsuperscript{\rm 4}
    \quad Haizhou Li\textsuperscript{\rm 1,3,5}
}
\affiliations {
    \textsuperscript{\rm 1}National University of Singapore, Singapore\\
    \textsuperscript{\rm 2}Peking University, China\\
    \textsuperscript{\rm 3}The Chinese University of Hong Kong, Shenzhen, China\\
    \textsuperscript{\rm 4}Tencent AI Lab, China\\
    \textsuperscript{\rm 5}Shenzhen Research Institute of Big Data, China\\
    \textsuperscript{\rm 6}University of Science and Technology of China, China\\

    chen\_zhang@u.nus.edu, jeffreyjiang@cuhk.edu.cn
}

\usepackage{bibentry}
\begin{document}

\maketitle

\begin{abstract}
In natural human-to-human conversations, participants often receive feedback signals from one another based on their follow-up reactions. These reactions can include verbal responses, facial expressions, changes in emotional state, and other non-verbal cues. Similarly, in human-machine interactions, the machine can leverage the user's follow-up utterances as feedback signals to assess whether it has appropriately addressed the user's request. Therefore, we propose using the likelihood of follow-up utterances as rewards to differentiate preferred responses from less favored ones, without relying on human or commercial LLM-based preference annotations. Our proposed reward mechanism, ``Follow-up Likelihood as Reward" (FLR), matches the performance of strong reward models trained on large-scale human or GPT-4 annotated data on 8 pairwise-preference and 4 rating-based benchmarks. Building upon the FLR mechanism, we propose to automatically mine preference data from the online generations of a base policy model. The preference data are subsequently used to boost the helpfulness of the base model through direct alignment from preference (DAP) methods, such as direct preference optimization (DPO). Lastly, we demonstrate that fine-tuning the language model that provides follow-up likelihood with natural language feedback significantly enhances FLR's performance on reward modeling benchmarks and effectiveness in aligning the base policy model's helpfulness\footnote{Repository is available at \url{https://github.com/e0397123/FLR}.}.

\end{abstract}


\section{Introduction}

The recent development of large language models (LLMs) has revolutionized the field of natural language processing~\citep{zhao2023survey}. Many chat-based LLMs are trained to align a large-scale pretrained language model to human preferences using techniques such as supervised fine-tuning (SFT)~\citep{wei2022finetuned,chung2022scaling}, reinforcement learning from human feedback (RLHF)~\citep{christiano-etal-2017-deep,ziegler2019finetuning,ouyang2022training}, or direct alignment from preferences (DAP)~\citep{rafailov2023direct,zhao2023slichf,ethayarajh2024kto}. All these techniques heavily rely on high-quality SFT or preference data, but collecting such data through human efforts is extremely expensive. 

\begin{figure}[!t]
\begin{subfigure}{0.49\linewidth}
  \centering
  % include first image
  \includegraphics[width=\textwidth]{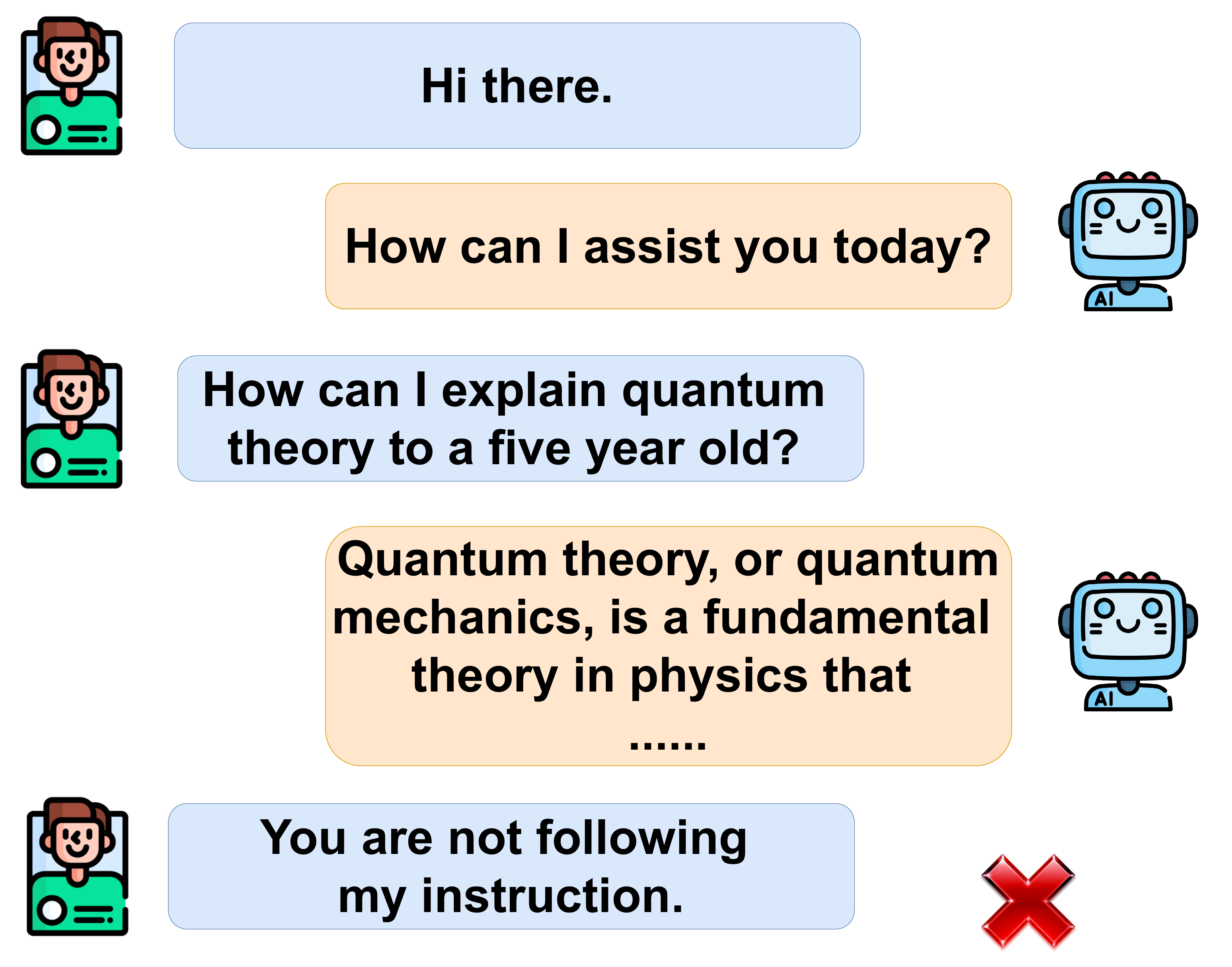} 
\end{subfigure}
\begin{subfigure}{0.49\linewidth}
  \centering
  % include second image
  \includegraphics[width=\textwidth]{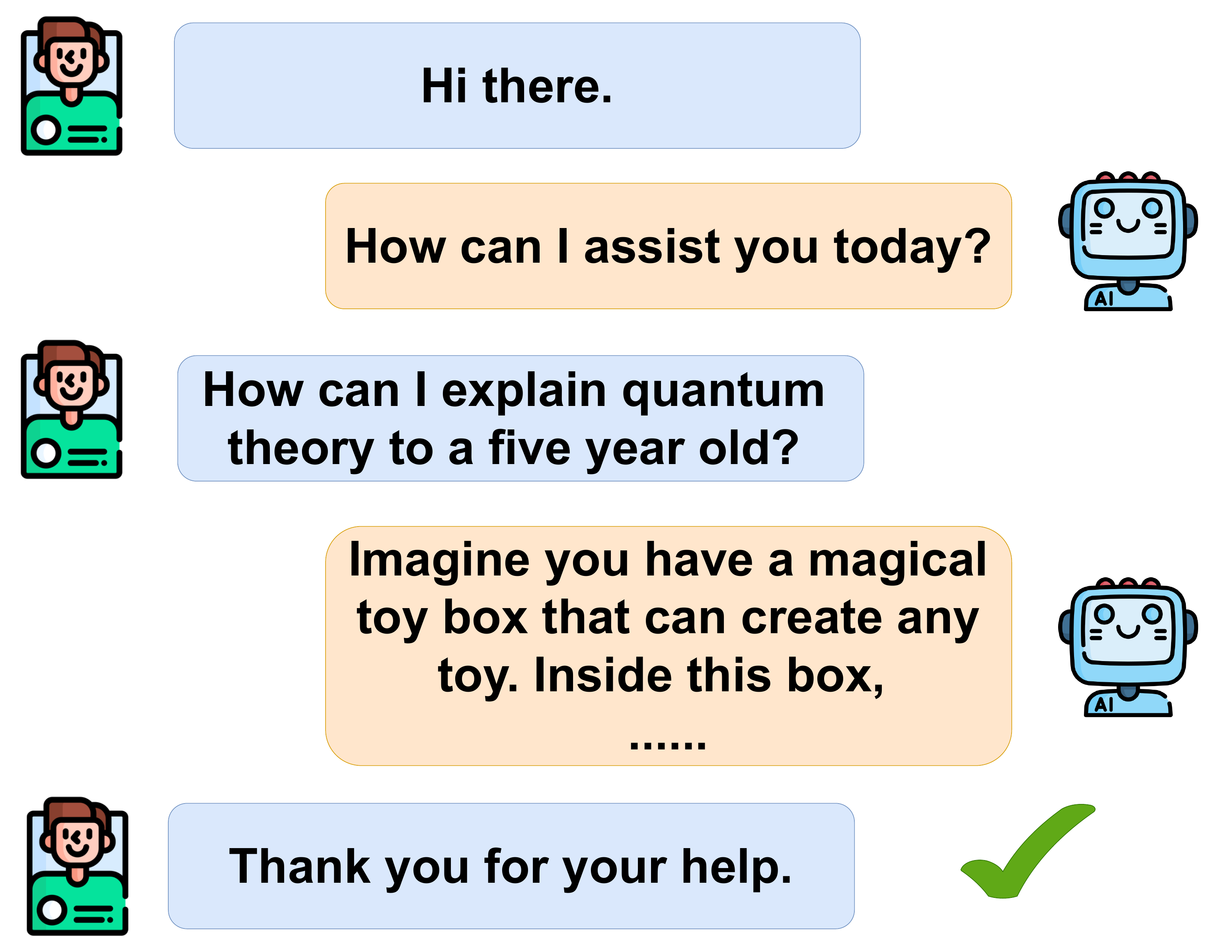}  
\end{subfigure}
\caption{Examples of a negative follow-up utterance (left) and a positive follow-up utterance (right) from the user .}
\label{fig:followup-example}
\end{figure}

Lately, there has been a growing trend in the self-evolution of LLMs. This process allows the models to autonomously obtain, improve, and learn from their own generated experiences~\citep{tao2024survey}, reducing the need for costly human supervision. For instance,~\citet{yuan2024selfrewarding} introduces self-rewarding language models, starting with bootstrapping instructions from a base policy model to generate candidate responses. The model then employs ``LLM-as-a-Judge" prompting~\citep{zheng2023judging} to evaluate its own outputs, creating a collection of self-curated preference data. The data are then used to align the language model through directed preference optimization~\citep{rafailov2023direct}. In a similar vein, many recent works~\citep[inter alia]{xu2023things,gulcehre2023reinforced,chen2024selfplay,guo2024direct,li2024selfalignment,zhang-etal-2024-ts} explore using online generations of language models for self-improvement in an iterative manner. At the core of these works is an automated procedure for response quality annotation, which utilizes either self or external reward signals to rank or score the response candidates~\citep{chen-etal-2024-unveiling}. The reliability of this procedure depends significantly on the reward model's language understanding and quality discrimination capabilities, which highly correlate with the model's scale and the quality of its training data~\citep{zhang-etal-2024-comprehensive}. For example, GPT-4~\citep{openai2023gpt4} emerges as an ideal annotator due to its large model size and high-quality instruction-tuning data. 

% \footnote{We refer to the model that provides reward signals as the reward model, whether it is self or external.} 

However, training a reward model similar to GPT-4 or using API services to prompt such a model is resource-intensive. Therefore, finding alternative indicators of response quality is necessary. Drawing inspiration from natural human-human interactions where speakers continuously adapt their speech based on real-time feedback from their conversational partners, we postulate that the quality of the LLMs' responses may be implicitly derived from the real users' reactions, their follow-up utterances in particular. For instance, if an AI assistant is not following the user's instructions, the users are more likely to express disagreement or criticism than compliments or gratitude. This is illustrated by the examples in Figure~\ref{fig:followup-example}.

Considering that it is infeasible to conduct large-scale quality annotation with an interactive setup between a real user and the language model, where the user provides real-time feedback to the language model generations, we approximate the user using an instruction-tuned LLM, which can either be the base policy model or another LM. This approach is motivated by the fact that existing instruction-tuned LLMs are fine-tuned using high-quality chat-style data~\citep{vicuna2023,xu-etal-2023-baize} and hence, possess the knowledge of human conversational dynamics. Consequently, their likelihood of generating negative feedback for an appropriate response is lower than that of generating positive feedback, and vice versa.

Hence, we manually curate a set of positive and negative follow-up utterances. The conditional log probability of these follow-up utterances, given a prompt-response pair, serves as a reward signal to determine the helpfulness of the response. While prior works~\citep{mehri-eskenazi-2020-unsupervised,de-bruyn-etal-2022-open} utilize the likelihood of follow-up utterances as an automatic evaluation metric for open-domain dialogues, our work takes a significant leap beyond mere dialogue evaluation. We leverage follow-up utterances for automatic preference annotations of online generations of the policy model without the need for an external reward model or human efforts. Such an automatic annotation procedure simplifies the alignment process and opens the door to more autonomous and efficient LLM optimization.

In summary, we make the following contributions:

\begin{itemize}

    \item We introduce a ``Follow-up Likelihood as Reward" (FLR) that automatically annotates preference data from base policy outputs. The annotated data is used for DAP fine-tuning, significantly enhancing the helpfulness of the base policy model. For example, Llama-3-8B-Instruct's length-controlled win rate is improved by 4.45\% on Alpaca-Eval V2 after fine-tuning (\S\ref{subsec:alignment-performance}).
    
    \item We demonstrate that FLR, without additional training, matches the performance of strong reward models, which are trained on large-scale human-annotated or GPT-4 annotated preference data, across 8 pairwise preference and 4 rating-based benchmarks (\S\ref{subsec:rm-results}).
    
    \item Lastly, we demonstrate that fine-tuning the LM used in FLR with natural language feedback data significantly enhances the performance on the reward modeling task (\S\ref{subsec:rm-results}) and results in better helpfulness alignment of the base policy LLM, compared to using the original FLR without fine-tuning (\S\ref{subsec:alignment-performance}).

\end{itemize}

\section{Related Work}

\subsubsection{Reward Model} Reward models play a pivotal role in the LLM alignment pipeline, serving as proxies for human judgment to guide policy models in distinguishing between desirable and undesirable outputs.~\citet{lambert2024rewardbench} categorizes reward models into three primary types: 1) classifier-based, 2) DPO-based, and 3) prompting or generation-based. The first category, exemplified by models such as Starling-RM-7B-alpha~\citep{starling2023}, Oasst RM, and ArmoRM-Llama3-8B-v0.1~\citep{wang2024interpretable}, involves training a sequence classifier on top of a language model using human-annotated preference data. The second category, including Tulu-2~\citep{wang2023how} and Zephyr-7b-alpha~\citep{tunstall2023zephyr}, directly computes rewards based on the conditional log probability of the response. The third category, such as auto-J~\citep{li2024generative}, Prometheus~\citep{kim2024prometheusv2}, and GPT-based LLMs~\citep{openai2023gpt4}, uses a prompting approach to generate reward scores, following the LLM-as-a-Judge~\citep{zheng2023judging} framework. The first category demands additional training efforts on large-scale human-annotated data to achieve satisfactory performance, while the third category typically requires a powerful, large-scale instruction-following LLM to attain similar levels of performance. Training such an LLM is exceedingly costly. Experiments in both~\citet{lambert2024rewardbench} and ours demonstrate the performance of the second category is sub-optimal compared to the first due to the lack of additional training.

However, FLR differs from all three categories. It shares the advantage of the second category by not requiring additional training on human-annotated data, yet it achieves performance comparable to that of the first category. Additionally, FLR paves the way for future research into leveraging feedback signals from real users in reward modeling. 

\subsubsection{LLM Alignment} Post-training alignment of large language models has become a key focus in LLM research. \citet{ouyang2022training} demonstrated the efficacy of the RLHF pipeline in enhancing GPT-3's ability to follow instructions. Subsequent studies have explored alternatives to costly reinforcement training by proposing techniques for direct alignment from preferences~\citep{rafailov2023direct,ethayarajh2024kto,zhao2023slichf}. Additionally, there has been significant interest in the self-improvement capabilities of LLMs~\citep{yuan2024selfrewarding,guo2024direct,wu2024selfplay} leveraging the DAP techniques and preference annotations on the online generation of the policy model. Our work focuses on reward modeling and the automatic annotation of preference data, making it applicable to all these algorithms. We also demonstrate the superiority of FLR in automatic preference annotation.

Recent studies have also explored the use of natural language feedback, produced by either humans or large language models (LLMs), to enhance the interpretability of LLM evaluations~\citep{cui2023ultrafeedback,mcaleese2024llm,madaan2023selfrefine,chen2024teaching,wu2024metarewarding}. This approach not only improves their utility but also aids in enhancing their helpfulness. Unlike existing methods, our work transforms this feedback into real-user follow-ups and utilizes it to refine the follow-up likelihood estimation in LLMs, thereby advancing the reward modeling capabilities of the proposed FLR mechanism.

\section{Methodology}
\label{sec:method}

In this section, we begin by formally defining the tasks. Next, we describe the ``Follow-up Likelihood as Reward (FLR)" procedure. Finally, we outline the automatic annotation process for the base language model's generations and explain how the data are used to align the base model.

\subsection{Task Formulation}
\label{subsec:formulation}

\subsubsection{Rating-Based Quality Annotation}

Given a chat-style instruction-tuning dataset $\{p_i, r_i\} \in \mathcal{D}$, $p_i$ is chat history between a user and a policy model and $r_i$ is the model completion. $p_i$ can be a single-turn instruction prompt from the user or a multi-turn interaction between the user and the policy model up to the most recent user turn. Based on certain criteria, we need to score $r_i$. Denote the score as $s_{r_i}$. In this paper, we focus on the helpfulness criteria of the completion as defined in~\citet{askell2021general}. We adopt an instruction-tuned LLM, denoted as $\mathcal{M}$ to conduct the FLR scoring mechanism. The effectiveness of FLR can be quantified by the agreement between $\{s_{r_i}\}_{i=1}^{|\mathcal{D}|}$ and the corresponding ground-truth ratings, $\{g_{r_i}\}_{i=1}^{|\mathcal{D}|}$. Common agreement metrics include Pearson, Spearman, and Kendall correlation coefficients. In most open-source instruction-tuning benchmarks~\citep{ye2024flask,kim2024prometheus,wang-etal-2024-helpsteer}, the ground-truth ratings are typically obtained via human evaluation or API-based prompting of GPT-4.

 % The common ways to score $r_i$ include human evaluation and model evaluation. As human evaluation or API-based LLM prompting approaches are expensive and non-scalable, especially for large-scale annotation of online generations of LM,

\subsubsection{Preference Data Annotation}

Most formulations of preference annotations is similar to rating-based quality annotation. The difference is that instead of a single completion $r_i$, there are $k\ge2$ candidates to score. The preference dataset, $\mathcal{D}$ consists of $\{p_i, r_i^1, r_i^2, \ldots, r_i^k\}$. Our goal is to rank $\{r_i^1, r_i^2, \ldots, r_i^k\}$ according to their extent of helpfulness. In this paper, we primarily study the special case of $k = 2$ and the effectiveness of the FLR scoring mechanism is quantified by the accuracy of scoring human preferred completion, $r_i^{+}$, higher than the disfavored completion, $r_i^{-}$, i.e., $s_{r_i^{+}} > s_{r_i^{-}}$.

% As most benchmarks~\citep{bai2022training,dubois2023alpacafarm,lambert2024rewardbench} and popular DAP algorithms adopt pairwise preference data, 

% Let $\pi_{rm}$ denote an instruction-tuned language model used for follow-up likelihood computation and $\mathcal{U} = \{u_1, u_2, u_3\}$ 

\subsection{Follow-up Likelihood as Reward (FLR)}
\label{subsec:flr-mechanism}

Instead of directly prompting LLMs~\citep{zheng2023judging} or explicitly training a reward model, we leverage natural follow-up utterances as implicit feedback to reflect the helpfulness of $r_i$. This approach is grounded in the observation that many powerful instruction-tuned LLMs~\citep{dubey2024llama,yang2024qwen2} are fine-tuned on high-quality conversational data that inherently reflect human-like feedback dynamics, whereby helpful responses are more likely to be followed by positive feedback, while unhelpful responses are more likely to trigger negative reactions. This procedure does not require additional preference data collection or the training of a separate reward model. It relies on the language-understanding capability of the LLMs.

Let us define a positive follow-up utterance as $f_j^+$ and a negative follow-up utterance as $f_j^{-}$. The log probabilities of the instruction-tuned LLM, $\mathcal{M}$ , generating $f_j^{+}$ and $f_j^{-}$ conditioned on $\{p_i, r_i\}$ are given by:

\begin{equation*}
    \log p_{\mathcal{M}}(f_j^* \mid p_i, r_i) = \sum_{t=1}^{T} \log p_{\mathcal{M}}(f_{j,t}^* \mid p_i, r_i, f_{j,<t}^*)
\end{equation*}
where $f_j^*$ can be either $f_j^+$ or $f_j^-$ and $T$ represents the total number of tokens in $f_j^*$.

% An ideal $\mathcal{M}$ satisfies the following constraints:

% \begin{align*}
%     &\log p_{\mathcal{M}}(f_j^+ \mid p_i, r_i^{+}) > \log p_{\mathcal{M}}(f_j^- \mid p_i, r_i^{+}) \quad (1) \\
%     &\log p_{\mathcal{M}}(f_j^+ \mid p_i, r_i^{-}) < \log p_{\mathcal{M}}(f_j^- \mid p_i, r_i^{-}) \quad (2) \\ 
%     & \log p_{\mathcal{M}}(f_j^+ \mid p_i, r_i^{+}) > \log p_{\mathcal{M}}(f_j^+ \mid p_i, r_i^{-}) \quad (3) \\
%     & \log p_{\mathcal{M}}(f_j^- \mid p_i, r_i^{+}) < \log p_{\mathcal{M}}(f_j^- \mid p_i, r_i^{-}) \quad (4)
% \end{align*}

Due to the large space of potential positive and negative follow-ups of $\{p_i, r_i\}$, relying on a single follow-up can introduce bias. Therefore, it is essential to consider a diverse set of follow-up utterances to accurately assess response quality. However, conducting an exhaustive search for all contextually relevant follow-ups is intractable. As a solution, we manually curate a set of positive and negative follow-ups based on the decomposition of helpfulness criteria. We leave the automatic search for suitable follow-ups to future work. Since helpfulness is an abstract concept, we decompose it into three fine-grained sub-categories: understanding, engagingness, and instruction-following. This decomposition is also motivated by prior works~\citep{wu2023finegrained,wang-etal-2024-helpsteer}. For each category, we come up with 10 different positive and negative follow-ups respectively. Table~\ref{tab:follow-up-examples} gives some examples and the full list is in Appendix A\footnote{The full appendix of the paper can be found at \url{https://arxiv.org/abs/2409.13948}.}.

\begin{table*}[!t]	
\centering
\resizebox{0.6\linewidth}{!}{
    \begin{tabular}{l|l}
    \toprule
    \textbf{Category}  & \textbf{Positive/Negative Follow-Up Example} \\ \midrule
    \multirow{2}{*}{Understanding} & That makes perfect sense! (\ding{51}) \\
    & That makes no sense! (\ding{55}) \\ \midrule
    % \multirow{2}{*}{Relevance} & That's directly related to my query. (\ding{51})  \\
    % & Your response isn't relevant to my query. (\ding{55}) \\ \midrule
    \multirow{2}{*}{Engagingness} & This is very interesting. (\ding{51}) \\
    & That’s not very interesting. (\ding{55}) \\ \midrule
    \multirow{2}{*}{Instruction-Following} & You did a fantastic job following my instructions. (\ding{51})\\
    & You didn't adhere to my instructions. (\ding{55}) \\
    % \multirow{2}{*}{Informativeness} & Your answer provided a lot of insight. ($\checkmark$) \\
    % & That’s a very generic response. (\ding{55}) \\ \midrule
    % \multirow{2}{*}{Correctness} & You gave me the correct details. (\ding{51})\\
    % & Your response is misleading. (\ding{55}) \\
    \bottomrule
    \end{tabular}
}
% 	\end{subtable}
\caption{Examples of positive and negative follow-ups for each sub-category of helpfulness.}
\label{tab:follow-up-examples}
\end{table*}

We define the set of positive follow-ups for a particular sub-category, $t$, as $F^{+}_{t}$ where $f_j^+ \in F^{+}_{t}$ and the set of negative follow-ups for a particular sub-category as $F^{-}_{t}$ where $f_j^- \in F^{-}_{t}$. To score $\{p_i, r_i\}$ based on the follow-ups of a particular sub-category, we define the following reward:

\begin{equation}
\resizebox{\linewidth}{!}{
    $s^t_{r_i} = \frac{\sum\limits_{f_j^+ \in F^{+}_{t}}\log p_{\mathcal{M}}(f_j^+ \mid p_i, r_i)}{|F^{+}_{t}|} - \frac{\sum\limits_{f_j^- \in F^{-}_{t}}\log p_{\mathcal{M}}(f_j^- \mid p_i, r_i)}{|F^{-}_{t}|}$
}
\label{eq:category-flr-score}
\end{equation}
The above reward signal is directly applied to solve the rating-based quality annotation task by average aggregation of scores across the three sub-categories\footnote{T = \{understanding, engagingness, instruction-following\}}. 

\begin{equation}
    s_{r_i} = \frac{\sum\limits_{t \in T}{s^t_{r_i}}}{|T|}
    \label{eq:average-flr-score}
\end{equation}

For the pairwise preference annotation task, an ideal $\mathcal{M}$ satisfies the following constraint: $\quad s^t_{r^+_i} > s^t_{r^-_i} \Rightarrow  s_{r^+_i} > s_{r^-_i}$.

% \resizebox{0.9\linewidth}{!}{%
%   \begin{minipage}{\linewidth}
% \begin{align*}
%     &\text{(a)} \quad \frac{\sum\limits_{f_j^+ \in F^{+}_{t}}\log p_{\mathcal{M}}(f_j^+ \mid p_i, r^+_i)}{|F^{+}_{t}|} > \frac{\sum\limits_{f_j^+ \in F^{+}_{t}}\log p_{\mathcal{M}}(f_j^+ \mid p_i, r^-_i)}{|F^{+}_{t}|}\\
%     &\text{(b)} \quad \frac{\sum\limits_{f_j^- \in F^{-}_{t}}\log p_{\mathcal{M}}(f_j^- \mid p_i, r^+_i)}{|F^{-}_{t}|} < \frac{\sum\limits_{f_j^- \in F^{-}_{t}}\log p_{\mathcal{M}}(f_j^- \mid p_i, r^-_i)}{|F^{-}_{t}|} \\
%     &\text{(c)} \quad s^t_{r^+_i} > s^t_{r^-_i} \Rightarrow  s_{r^+_i} > s_{r^-_i} \\
% \end{align*}
%   \end{minipage}  
% }

It is worth noting that FLR can seamlessly integrate into the autonomous self-evolution pipeline of language models via online DAP procedures (details in \S\ref{subsec:alingn-base-lm}). it does not rely on collecting offline human preference data or external model supervision but instead leverages the internal knowledge of strong instruction-tuned models. 

\subsubsection{Enhance FLR with Natural Feedback Data}

To enhance the ability of $\mathcal{M}$ in assigning high likelihoods to positive follow-ups and low likelihoods to negative follow-ups for helpful responses, and vice versa for unhelpful responses, we finetune $\mathcal{M}$ using natural language feedback data using a chat template as shown below  

\begin{promptbox}
\textbf{User:} Please ask me a question or assign me a task.
\end{promptbox}

\begin{responsebox}
\textbf{LLM:} [insert here the instruction prompt, $p_i$]
\end{responsebox}

\begin{promptbox}
\textbf{User:} [insert here the response, $r_i$]
\end{promptbox}

\begin{responsebox}
\textbf{LLM:} [insert here the feedback to $(p_i, r_i)$, which is denoted as $l_i$]
\end{responsebox}

In the chat template, the roles of the LLM and user are switched. A feedback data example can be found in Appendix B. During the supervised fine-tuning of $\mathcal{M}$, the model is optimized to generate the natural language feedback, $l_i$. To curate training data, a common approach is to mine it from real user-bot interactions. However, this real-world data often contains significant noise and demands costly cleaning efforts. As an alternative, we rewrite feedback from third-party evaluators in an existing dataset, transforming it into naturally occurring utterances with a conversational, first-person tone. Specifically, we prompt GPT-3.5-Turbo to rewrite the feedback sentences in a first-person tone. This strategy aligns the feedback more closely with real-user follow-ups. The instruction template to prompt ChatGPT is presented in Appendix C.

\subsection{Aligning Base Language Model}
\label{subsec:alingn-base-lm}

\begin{figure}[!t]
  \centering
  % include first image
  \includegraphics[width=\linewidth]{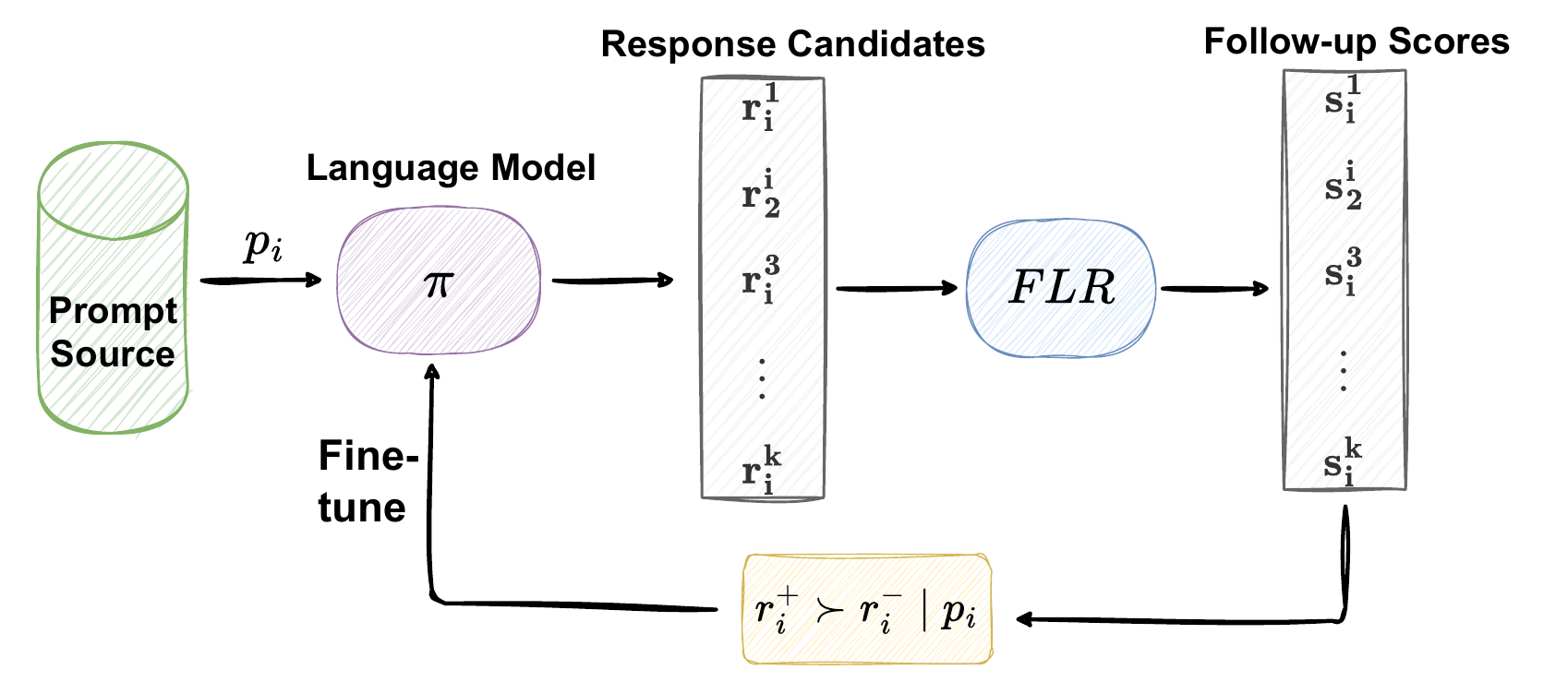} 
\caption{The procedure of aligning base LM. FLR denotes the ``follow-up likelihood as reward" mechanism.}
\label{fig:dap-process}
\end{figure}

Figure~\ref{fig:dap-process} presents the overall procedure of aligning a base policy model $\pi$. The main idea is to first sample instruction prompts $p_i$ from a prompt source and $\pi$ generates $k$ response candidates, $\{r_i^1, r_i^2, \ldots, r_i^k\}$. Each candidate is assigned a reward score as defined in Eq.\ref{eq:average-flr-score}. The candidate with the highest score is treated as the positive response $r^+_i$. The candidate with the lowest score is treated as the negative response $r^-_i$. This strategy helps avoid using similar pairs with slight differences in reward scores, which may confuse the generation model. Additionally, if the obtained pair contains responses with the same reward scores, it is filtered out. We experiment with DPO~\citep{rafailov2023direct} and KTO~\citep{ethayarajh2024kto} fine-tuning of $p_i$ using the synthetic preference dataset constructed with the above procedure.

% We experiment with SFT,  DPO~\citep{rafailov2023direct}, and KTO~\citep{ethayarajh2024kto} fine-tuning of $p_i$ using the synthetic preference dataset constructed using the above procedure. 

% For SFT fine-tuning, we only need the positive response $r^+_i$ for the instruction prompt $p_i$.

\section{Experiment Setup}
\label{sec:experiment-setup}

There are two parts of our experiments: (1) reward modeling and (2) helpfulness alignment. For (1), reward models are assessed on human-annotated pairwise preference and rating-based quality evaluation benchmarks (both tasks have been introduced in \S2.1). (2) involves evaluating the helpfulness alignment of policy models after fine-tuning using the automatically curated preference data with FLR. \S3.1 presents the training details. \S3.2 introduces benchmarks and evaluation methodology used in both parts of the experiments. \S3.3 describes the baselines for comparison.

\begin{table*}[!t]
\centering
\resizebox{0.7\linewidth}{!}{
\begin{tabular}{l|c|c|c|c}
\toprule
\textbf{Test Datasets} & \textbf{Size} & \makecell{\textbf{Avg. \#Prompt} \\ \textbf{Words}}  & \makecell{\textbf{Avg. \#Turns} \\ \textbf{per Prompt}} & \textbf{Type} \\ \midrule
HH-RLHF Helpfulness~\shortcite{bai2022training} (HH) & 6,238 & 93.05 & 2.38 & Pairwise \\
BeaverTails Helpfulness~\shortcite{ji2024beavertails} (BH) & 2,985 & 13.17 & 1.00 & Pairwise \\
SHP~\shortcite{pmlr-v162-ethayarajh22a} & 18,409 & 148.79 & 1.00 & Pairwise \\
Alpaca-Farm~\shortcite{dubois2023alpacafarm} (AF) & 17,701 & 28.57 & 1.00 & Pairwise \\
MT-Bench Pairwise~\shortcite{zheng2023judging} (MP) & 916  & 235.73 & 3.00 & Pairwise \\
RewardBench Chat Easy~\shortcite{lambert2024rewardbench} (RCE) & 358 & 41.92 & 1.00 & Pairwise \\
RewardBench Chat Hard~\shortcite{lambert2024rewardbench} (RCH) & 456 & 30.45 & 1.00 & Pairwise \\
Preference Bench~\shortcite{kim2024prometheus} (PB) & 1,998 & 79.21 & 1.00 & Pairwise \\
\midrule
Feedback Bench~\shortcite{kim2024prometheus} (FH) & 1,000 & 79.25 & 1.00 & Rating \\
FLASK~\shortcite{ye2024flask} & 1,500 & 70.29 & 1.00 & Rating \\
MT-Bench Rating~\shortcite{zheng2023judging} (MR) & 1,000 & 141.97 & 2.05  & Rating \\
HelpSteer~\shortcite{wang-etal-2024-helpsteer} & 1,789 & 432.42 & 1.00 & Rating \\
\bottomrule
\end{tabular}
}
\caption{Statistics of the reward model benchmarks.}\label{tab:rm-benchmarks}
\end{table*}

\subsection{Training Details}

For $\mathcal{M}$, we experiment with Llama-3-8B-Instruct and Qwen2-7B-Instruct. We utilize the UltraFeedback~\citep{cui2023ultrafeedback} dataset for rewriting, which is then employed to fine-tune $\mathcal{M}$. UltraFeedback contains 255,548 instances of ($p_i$, $r_i$, $l_i$) and we randomly sample 100K for fine-tuning. 

For the helpfulness alignment experiment, we also adopt Llama-3-8B-Instruct and Qwen2-7B-Instruct as the base policy model $\pi$. LoRA~\citep{hu2022lora} is applied to all fine-tuning experiments on a single NVIDIA A100 80GB GPU. The experimental settings for DPO and KTO fine-tuning follow the implementations from LLaMA-Factory~\citep{zheng2024llamafactory}. We adopt the Nectar dataset~\citep{starling2023} as the prompt source. The 183K prompts in Nectar are a mixture of diverse sources, including lmsys-chat-1M~\citep{zheng2023lmsyschat1m}, ShareGPT, Anthropic/HH-RLHF~\citep{bai2022training}, UltraFeedback, Evol-Instruct~\citep{xu2023wizardlm}, and Flan~\citep{longpre2023flan}. For our experiments, we randomly sampled 100K instruction prompts from Nectar. Full details on reproducibitlity can be found in Appendix E.

\begin{table*}[!ht]
\centering
\resizebox{\linewidth}{!}{
\begin{tabular}{l|cccccccc|c|cccc|c}
\toprule
\textbf{Models} & \textbf{HH} & \textbf{BH} & \textbf{SHP} & \textbf{AF} & \textbf{MP} & \textbf{RCE} & \textbf{RCH} & \textbf{PB} & \textbf{AVG-P} & \textbf{MR} & \textbf{FLASK} &\textbf{FB} & \textbf{HelpSteer} & \textbf{AVG-R}  \\ \midrule
Direct (Tulu-2-dpo-7b) & 55.59 & 58.99 & 45.83 & 55.90 & 63.65 & 86.59 & 39.04 & 62.51 & 58.51 & 0.114 & 0.111  & 0.206 & 0.167 & 0.150 \\
Direct (Qwen2-7B-Instruct) & 54.15 & 52.86 & 40.81 & 54.77 & 63.43 & 86.03 & 47.37 & 54.50 & 56.74 & 0.184 & 0.117 & 0.050 & 0.077 & 0.107 \\
Direct (Zephyr-7b-alpha) & 53.01 & 54.17 & 40.49 & 56.29 & 61.35 & 80.17 & 53.73 & 61.41 & 57.58 & 0.191 & 0.004 & 0.201 & 0.057 & 0.113 \\
Direct (Llama-3-8B-Instruct) & 55.51 & 54.74 & 42.64 & 54.90 & 62.88 & 86.31 & 43.86 & 49.25 & 56.26 & 0.183 & 0.089 & 0.024 & 0.127 & 0.106 \\ \midrule
GPT-3.5-Turbo & 59.03 & 63.87 & 61.17 & 59.70 & 71.89 & 82.26 & 43.09 & 84.18 & 65.65 & 0.439 & 0.016 & 0.023 & 0.025 & 0.126 \\
Prometheus-7B & \textbf{62.95} & \textbf{66.03} & 55.13 & \textbf{64.01} & \textbf{80.35} & 92.18 & 48.03 & \textbf{95.45} & \textbf{70.52} & 0.312 & 0.251 & \textbf{0.871} & 0.390 & 0.456 \\
Oasst-Pythia-6.9B & 62.18 & 60.67 & \textbf{68.61} & 56.21 & 72.93 & 91.9 & 37.72 & 77.73 & 65.99 & 0.284 & 0.137 & 0.526 & 0.329 & 0.319  \\
PairRM & 62.06 & 56.82 & 54.91 & 58.04 & 76.09 & 84.64 & 50.66 & 81.03 & 65.53 & - & - & -& - & -  \\
Starling-RM-7B-alpha & \textbf{63.11} & \textbf{71.69} & 59.67 & \textbf{60.20} & \textbf{78.38} & \textbf{94.13} & 40.35 &86.69 & 69.28 & 0.492 & 0.235 & \textbf{0.762} & 0.449 & 0.485 \\
ArmoRM-Llama3-8B-v0.1 & - & 63.72 & \textbf{68.87} & \textbf{60.04} & 77.40 & \textbf{97.21} & \textbf{76.54} & \textbf{90.59} & \textbf{76.34} & 0.406 & \textbf{0.348} & \textbf{0.778} & \textbf{0.452} & \textbf{0.496} \\ \midrule
FLR (Llama-3-8B-Instruct) & 58.66 & 55.71 & 57.79 & 56.07 & 74.56 & 90.50 & 63.16 & 75.03 & 66.44 & \textbf{0.555} & 0.285 & 0.509 &  0.383 & 0.433 \\ 
FLR (Qwen2-7B-Instruct) & 57.23 & 58.29 & 59.37 & 53.39 & 71.62 & 80.45 & 55.26 & 73.12 & 63.59 & 0.403 & 0.157 & 0.423 & 0.296 & 0.320 \\ 
FLR-FT (Llama-3-8B-Instruct) & \textbf{62.92} & \textbf{65.09} & \textbf{61.19} & 59.99 & \textbf{79.91} & \textbf{95.53} & \textbf{55.92} & \textbf{87.64} & \textbf{71.02} & \textbf{0.606} &  \textbf{0.430} & 0.705 & \textbf{0.566} & \textbf{0.577} \\ 
FLR-FT (Qwen2-7B-Instruct) & 62.50 & 61.11 & 57.17 & 59.28 & 72.93 & 92.18 & \textbf{69.30} & 85.39 & 69.98 & \textbf{0.571} &  \textbf{0.504} & 0.701 & \textbf{0.524} & \textbf{0.575} \\ 
\bottomrule
\end{tabular}
}
\caption{Results on the pairwise preference and rating-based helpfulness benchmarks. AVG-P (\%) and AVG-R denote the average scores of pairwise preference and rating-based benchmarks respectively. The HH score for ArmoRM is omitted as it is trained on the data, achieving nearly 100\% accuracy. FT refers to fine-tuning with the natural language feedback. Top-3 scores for each benchmark are highlighted in bold. ``Direct" denotes direct scoring of the response by the LM using the log probability.}\label{tab:rm-main-results}
\end{table*}

\subsection{Benchmarks and Evaluation}

We assess the reward models using eight pairwise preference benchmarks and four rating-based single-response benchmarks, with their statistics in Table~\ref{tab:rm-benchmarks}. Accuracy is reported for the pairwise benchmarks and Pearson correlation for the rating-based benchmarks. Additionally, we examine the reward models' ability to rank different LLMs by providing system-level correlations between the reward model scores and real-user ELO ratings from the LMSys Chatbot Arena~\citep{chiang2024chatbot}. We obtain the Arena dataset\footnote{The ELO ratings are updated as of June 7, 2024.} from~\citet{lin2024wildbench} and it consists of 865 data instances per LLM for a total of 30 LLMs, including GPT-4-turbo~\citep{openai2023gpt4}, Meta-Llama-3-70B-Instruct, and Mixtral-8x7B-Instruct~\citep{jiang2024mixtral}.

To evaluate FLR's contribution to helpfulness alignment, we use well-established benchmarks including Alpaca-Eval V2~\citep{alpaca_eval}, WildBench V2~\citep{lin2024wildbench}, and FLASK~\citep{ye2024flask}.
Alpaca-Eval V2, WildBench V2, and FLASK contain 805, 1024, and 1700 instruction prompts respectively. WildBench V2 involves pairwise comparisons with the base policy model as the reference. For Alpaca-Eval V2, we report the length-controlled win rate against GPT-4 Preview (11/06) as per standard protocol, and the ``alpaca\_eval\_gpt4\_turbo\_fn" annotator config is adopted. For FLASK, the GPT-4 evaluator is adopted to rate the quality of the model responses according to specific prompts. Furthermore, we assess the general understanding capability of the aligned models on the Open LLM benchmark. We follow the official implementations of the benchmarks and the fp16 model variants are used for inference. 

% Lastly, we also conduct human evaluations of the alignment performance as detailed in \S\ref{subsec:alignment-performance}.

% Specifically, we sample 150 prompts from Alpaca-Eval V2, and ask humans to conduct pairwise comparison of the responses generated by the base policy model and the policy model after fine-tuning using the FLR-annotated preference data.

\subsection{Baselines}

Our proposed FLR mechanism is compared to state-of-the-art reward models, including Prometheus-7b-v2.0~\citep{kim2024prometheusv2}, GPT-3.5-Turbo, Starling-RM-7B-alpha~\citep{starling2023}, oasst-rm-2-pythia-6.9b-epoch-1, PairRM~\citep{jiang-etal-2023-llm}, and ArmoRM-Llama3-8B-v0.1~\citep{wang2024interpretable}. It’s important to note that these RMs are trained on a diverse mix of high-quality, human-annotated, or GPT-annotated preference data, whereas FLR operates without such annotations. As PairRM is trained only for preference annotations, its performance on rating-based benchmarks is not reported.

Moreover, FLR is compared to the direct scoring of responses based on their log probability, as judged by DPO-finetuned LMs. These include Tulu-2-dpo-7b~\citep{wang2023how}, Qwen2-7B-Instruct~\citep{yang2024qwen2}, Zephyr-7b-alpha~\citep{tunstall2023zephyr}, and Llama-3-8B-Instruct. We follow the official implementation of RewardBench~\citep{lambert2024rewardbench} and prometheus-eval~\citep{kim2024prometheus}. 

For the alignment experiment, we primarily study how the alignment performance changes with respect to the base policy model. Note that we focus on models at the 7B scale due to their good balance between performance and computational resource requirements.

% \textcolor{red}{To ADD - We also compare the aligned language model with other strong instruction-tuned LLMs, LMs aligned with strong RMs.} 

\section{Experiment}

This section presents the results and analysis of the reward modeling and alignment experiments.

\subsection{Reward Modeling Results}
\label{subsec:rm-results}

\subsubsection{Main Benchmark Results} 

Table~\ref{tab:rm-main-results} presents the accuracy scores (\%) and Pearson correlations of various reward models on 8 pairwise preference datasets and 4 rating-based helpfulness datasets respectively. First, we observe that FLR performs significantly better than directly using the response likelihood. For instance, the average pairwise preference accuracy of FLR (Llama-3-8B-Instruct) is 66.44\%, compared to 56.26\% for Direct (Llama-3-8B-Instruct), showing a roughly 10\% difference. Additionally, the average Pearson correlation achieved by FLR (Llama-3-8B-Instruct) is 0.327 points higher than that of Direct (Llama-3-8B-Instruct). We can also observe that FLR (Llama-3-8B-Instruct) without fine-tuning outperforms GPT-3.5-Turbo, PairRM, and Oasst-rm-2-pythia-6.9b-epoch-1, three strong lightweight reward models. These observations demonstrate that using real user follow-ups as an indication of response quality is a promising direction of reward modeling, given the presence of a strong instruction-tuned LM. Even compared to Prometheus-7b-v2.0, Starling-RM-7B-alpha, and ArmoRM-Llama3-8B-v0.1, which are trained on large-scale and high-quality human-annotated or GPT-4-annotated preference data, the performance of FLR without fine-tuning looks promising, especially on the rating-based benchmarks. 

After fine-tuning with the natural language feedback data, the performance of FLR improves significantly. FLR (Llama-3-8B-Instruct - FT) achieves an average accuracy gain of 4.58\% on the pairwise preference datasets and an increase of 0.144 in average Pearson correlation on the rating-based benchmarks compared to FLR (Llama-3-8B-Instruct). Significant performance boost can also be observed when using Qwen2-7B-Instruct. Moreover, the FLR fine-tuned variants' reward modeling performance consistently ranks among the top three of all reward models on most of the benchmarks. On average, FLR (Llama-3-8B-Instruct - FT) ranks third on pairwise preference benchmarks and first on rating-based benchmarks. 

It's important to note that while ArmoRM-Llama3-8B-v0.1 ranks first on pairwise preference benchmarks, it has been extensively trained on around 1 million annotated pairs from 10 diverse, high-quality human preference datasets. Similarly, Prometheus-7b-v2.0 has also been trained on 20K preference data and 20K single-rating data, both with high-quality annotations. In contrast, vanilla FLR operates without such data, and its fine-tuning with approximately 10K natural language feedback samples isn't directly optimized for pairwise preference reward modeling. Moreover, the feedback data can be curated from real user-bot interactions or evaluation logs from third-party evaluators, showcasing the potential of using more accessible and naturally occurring data sources for effective reward modeling, bypassing the need for costly human or commercial LLM annotations.

\begin{table}[!t]
\centering
\resizebox{\linewidth}{!}{
\begin{tabular}{l|c|c|c}
\toprule
\textbf{Models} & \textbf{Pearson} & \textbf{Spearman} & \textbf{Kendall Tau} \\ \midrule
Prometheus-7B & 0.683	& 0.673	& 0.492 \\
Oasst-Pythia-6.9B & 0.755	& 0.702	& 0.533 \\
Starling-RM-7B-alpha & 0.849 & 0.850 & 0.658 \\
ArmoRM-Llama3-8B-v0.1 & \textbf{0.909} &	\textbf{0.897} & \textbf{0.751} \\
FLR (Llama3-8B-Instruct) & \textbf{0.897} & \textbf{0.899} & \textbf{0.727} \\
FLR (Qwen2-7B-Instruct) & 0.631 & 0.586 & 0.454 \\
FLR-FT (Llama3-8B-Instruct) & 0.860 & 0.895 & \textbf{0.727} \\
FLR-FT (Qwen2-7B-Instruct) & \textbf{0.892} & \textbf{0.897} & \textbf{0.732} \\
\bottomrule
\end{tabular}
}
\caption{System-level correlations on Chatbot Arena.}\label{tab:arena-system-corr}
\end{table}

\subsubsection{Correlation with Chatbot Arena}

In addition to ranking or rating responses, reward models can efficiently evaluate and rank the performance of different models. This capability provides researchers with quick feedback on model performance, facilitating faster model development and iteration. We examine whether FLR benefits such a use case. Table~\ref{tab:arena-system-corr} presents the system-level correlations of the reward models with real-user ELO ratings on LMSys Chatbot Arena. We can see that FLR (Llama-3-8B-Instruct), FLR (Llama-3-8B-Instruct - FT), and FLR (Qwen2-7B-Instruct - FT) achieve strong performance, comparable to the state-of-the-art reward model, ArmoRM-Llama3-8B-v0.1. The observations further reinforce the effectiveness of the FLR.

% Moreover, we observe no significant performance change in FLR using Meta-Llama-3-8B-Instruct before and after fine-tuning while fine-tuning significantly boosts the correlations of FLR using Qwen2-7B-Instruct. 

\begin{table}[!t]
\centering
\resizebox{0.9\linewidth}{!}{
\begin{tabular}{l|c|c}
\toprule
\textbf{Models} & Avg-P & Avg-R \\ \midrule
FLR (Llama-3-8B-Instruct) & 66.44 & 0.433 \\
\quad$\setminus$ Engagingness & 64.57 & 0.412 \\
\quad$\setminus$ Understanding & 66.27  & 0.405 \\
\quad$\setminus$ Following & 64.56 & 0.398 \\
\quad$\setminus$ Positive Follow-ups & 57.50 & -0.118 \\
\quad$\setminus$ Negative Follow-ups & 54.73 & 0.169 \\\midrule
FLR (Llama-3-8B-Instruct - FT) & 71.02 & 0.577 \\
\quad$\setminus$ Engagingness & 70.22 & 0.573 \\
\quad$\setminus$ Understanding & 70.75  & 0.552  \\
\quad$\setminus$ Following & 69.92 & 0.512 \\
\quad$\setminus$ Positive Follow-ups & 56.27 & -0.034 \\
\quad$\setminus$ Negative Follow-ups & 67.09 & 0.484 \\
\bottomrule
\end{tabular}
}
\caption{Ablation study for the reward modeling task. The symbol ``$\setminus$" denotes exclusion.}\label{tab:ablation-results}
\end{table}

\subsection{FLR Ablation Study}

As shown in Table~\ref{tab:ablation-results}, removing any of the three categories: engagingness, understanding, or instruction-following, leads to a performance reduction. This effect is observed in both Llama-3-8B-Instruct and Llama-3-8B-Instruct - FT. Removing the instruction-following category results in the most significant performance reduction, indicating that it is a crucial attribute of response helpfulness.

Additionally, we observe that using only positive follow-ups or only negative follow-ups significantly reduces performance. Using only positive follow-ups means retaining only the first term in Eq.\ref{eq:category-flr-score}, while using only negative follow-ups involves leveraging only the second term in Eq.\ref{eq:category-flr-score}. Ideally, FLR should maximize the first term and minimize the second term when evaluating a good response, whereas a poor response would exhibit the opposite characteristics. Our proposal of using the score differences between positive and negative follow-ups is more robust and demonstrates significantly better performance.

\subsection{Alignment Performance}
\label{subsec:alignment-performance}

\subsubsection{Main Results}
Table~\ref{tab:alignment-performance} presents the alignment performance of policy models fine-tuned with DPO or KTO using preference data annotated with our proposed FLR mechanism, compared to the un-finetuned models\footnote{FLASK evaluates 12 quality aspects of a response, we report the average score across all aspects here. Details can be found in Table 12 of Appendix D.}. Remarkably, we can observe that Llama-3-8B-Instruct can self-improve using our FLR mechanism without external supervision. For example, Llama-3-8B-Instruct + FLR DPO achieves a 33.17\% LC win rate on Alpaca-Eval V2, marking a 4.45\% improvement. On WildBench V2, it records a 57.57\% win rate over its base version. Meanwhile, Qwen2-7B-Instruct shows less significant self-improvement, with minor gains observed only on WildBench V2 and FLASK. This may be because FLR is sensitive to factors such as the training method and the type of training data used in the underlying language model. This insight further motivates enhancing FLR by fine-tuning the language model with natural language feedback data.

It can be observed that the fine-tuned LM delivers more accurate FLR rewards, i.e., better indication of response helpfulness. With FLR-FT DPO, both Llama-3-8B-Instruct and Qwen2-7B-Instruct demonstrate significant improvements across all benchmarks. Notably, Qwen2-7B-Instruct + FLR DPO achieves the highest LC win rate of 34.26\% on Alpaca-Eval V2 and a rating of 3.95 out of 5 on FLASK. 

We can also observe that DPO fine-tuning yields superior alignment performance compared to KTO, consistent with previous findings~\citep{rasul-etal-2024-preference} that DPO outperforms KTO under paired preference settings. However, KTO's flexibility lies in its ability to function without paired data, making it more versatile than DPO. It is worth noting that FLR is compatible with all the DAP algorithms.

\begin{table}[!b]
\centering
\resizebox{\linewidth}{!}{
\begin{tabular}{l|c|c|c|c}
\toprule
\textbf{Models} & FLASK & AlpacaEval LC & WildBench & Open LLM \\ \midrule
Llama-3-8B-Instruct & 3.76	& 28.72 & 50.00 & 0.644 \\
+ FLR DPO	& 3.72	& 33.17 & 57.57 & 0.648 \\
+ FLR KTO 	& 3.73	& 30.59 & 55.50 & 0.640 \\
+ FLR-FT DPO & \textbf{3.83} & \textbf{34.22} & \textbf{63.62} & \textbf{0.652} \\
+ FLR-FT KTO & 3.76	& 33.89 & 60.50 & 0.643 \\ \midrule 
Qwen2-7B-Instruct& 3.87 & 29.74 & 50.00 & 0.639 \\
+ FLR DPO 	& 3.92	& 28.78 & 55.57 & 0.632 \\
+ FLR KTO	& 3.89	& 26.15 & 54.49 & 0.640 \\
+ FLR-FT DPO & \textbf{3.95} & \textbf{34.26} & 61.52 & \textbf{0.657} \\
+ FLR-FT KTO  & 3.92 & 33.17 & \textbf{63.93} & 0.643 \\ 
\bottomrule
\end{tabular}
}
\caption{Alignment performance. The last column is the average score on the Open LLM Leaderboard. LC refers to the length-controlled win rate. FLR-FT refers to using the fine-tuned Llama-3-8B-Instruct to generate follow-up likelihood while FLR refers to using the base LM without fine-tuning.}\label{tab:alignment-performance}
\end{table}

\begin{table*}[!ht]
\centering
\resizebox{\linewidth}{!}{
\begin{tabular}{l|c|cccccc}
\toprule
\textbf{Methods} & \textbf{LC / Win Rate / Length} & \textbf{RM} & \textbf{\#External Anno} & \textbf{Reward Signal} & \textbf{Anno Type} & \textbf{Prompt Source} \\ \midrule

PairRM &  35.92 / 33.04 / 1847 & PairRM & $\sim$490K & Human, GPT-4 & Pairwise & UltraFeedback \\
ArmoRM & 41.40 / 39.13 / 1904 &  ArmoRM & $\sim$1M & Human, GPT-4 & Pairwise & UltraFeedback \\
FLR (UltraFeedback) & 32.60 / 34.22 / 2078 & FLR & 0 & Follow-ups & - & UltraFeedback \\
FLR-FT (UltraFeedback) & 35.18 / 35.47 / 2003 & FLR-FT &  100K & Follow-ups & NL Feedback & UltraFeedback \\
FLR (Nectar) & 33.17 / 35.16 / 2091  & FLR & 0 & Follow-ups & - & Nectar-100K \\
FLR-FT (Nectar) & 34.22 / 35.22 / 2049  & FLR-FT & 100K & Follow-ups & NL Feedback & Nectar-100K \\
\bottomrule
\end{tabular}
}
\caption{Comparative analysis on Alpaca-Eval V2. ``LC", ``FT", ``Anno", and ``NL" denote ``length-controlled", ``fine-tuned", ``annotations", and ``natural language" respectively. Results of the LLaMA-3-8B-Instruct base policy is provided for reference.}\label{tab:comparative-analysis}
\end{table*}

\subsubsection{General Understanding Capability} We also evaluate whether fine-tuning policy models using preference data annotated with FLR enhances response helpfulness without compromising their general understanding capability. The average scores of different models on the Open LLM Leaderboard are presented in Table~\ref{tab:alignment-performance}. The detailed results for each dataset in the leaderboard are presented in Table 13 of Appendix D. We observe that FLR does not compromise model performance on the Open LLM leaderboard; in fact, it enhances it. For instance, Llama-3-8B-Instruct + FLR DPO achieves an average score of 0.648 compared to 0.644 for the base model. FLR-FT DPO delivers the best performance, with scores of 0.652 and 0.657 for the respective Llama-3-8B-Instruct and Qwen2-7B-Instruct base models.

\subsubsection{Comparative Analysis}

Table~\ref{tab:comparative-analysis} and Table~\ref{tab:arena-hard-comparative-analysis} present the results of various online DPO pipelines on Alpaca-Eval V2, Arena-Hard~\citep{li2024crowdsourced}, and FLASK respectively. Table~\ref{tab:comparative-analysis} also highlights the key differences among them. We apply the same experiment recipe across pipelines, including Llama-3-8B-Instruct as the base policy model, Lora as the fine-tuning approach, DPO as the alignment technique, and the same set of training hyperparameters (details shown in Appendix E). We observe a performance gap between the state-of-the-art reward model, ArmoRM-Llama3-8B-v0.1, and our proposed FLR mechanism. ArmoRM achieves a 41.40\% LC win rate on Alpaca-Eval V2 and 28.4\% win rate on Arena-Hard respectively, whereas the best-performing FLR variant achieves 35.18\% and 24.8\% on the two benchmarks respectively. The gap can be attributed to the significant differences in both the quantity and diversity of annotated data utilized. ArmoRM, a Mixture-of-Experts (MoE) model, is trained on approximately 1 million pairwise annotations from humans or GPT-4, optimized across 19 reward objectives. In contrast, FLR bypasses the need for annotated preference data by deriving reward signals from naturally occurring follow-ups. 

Notably, FLR performs comparably to PairRM, which has been trained on approximately 490K high-quality annotations, on all three benchmarks. This underscores the potential of leveraging naturally occurring follow-ups as a reward signal. Building on FLR's promise, our results further demonstrate that fine-tuning with natural language (NL) feedback can enhance alignment performance between FLR and FLR-FT, though the increase in LC win rate remains modest. Future work could explore more effective ways to leverage NL feedback or focus on curating higher-quality NL feedback to further enhance FLR's performance. 

Additionally, we also examined the impact of using different prompt sources in the alignment pipeline and found no significant difference between using UltraFeedback and Nectar, such as 32.60\% vs 33.17\% LC win rate for FLR on Alpaca-Eval,  22.2\% vs 21.7\% win rate on Arena-Hard, and 3.76 vs 3.72 on FLASK. The observation is potentially due to the similarity in the distribution of their data.

\begin{table}[!t]
\centering
\resizebox{\linewidth}{!}{
\begin{tabular}{l|cc|c}
\toprule
\textbf{Methods} & \textbf{Arena-Hard} & \textbf{95\% CI} & \textbf{FLASK} \\ \midrule
LLaMA-3-8B-Instruct &  20.6 & (-2.0, 1.9) & 3.76 \\
PairRM &  25.2 & (-2.1, 2.0) & 3.81 \\
ArmoRM & 28.4 &  (-1.9, 1.9) & 3.81 \\
FLR (UltraFeedback) & 22.2 & (-2.0, 1.9) & 3.76  \\
FLR-FT (UltraFeedback) & 24.5 & (-1.7, 1.5) & 3.83 \\
FLR + (Nectar) & 21.7  & (-2.1, 2.4) & 3.72 \\
FLR-FT + (Nectar) & 24.8  & (-2.6, 2.3) & 3.83 \\
\bottomrule
\end{tabular}
}
\caption{Comparative analysis on Arena-Hard and FLASK.}\label{tab:arena-hard-comparative-analysis}
\end{table}

\section{Conclusion}

In conclusion, our proposed 'Follow-up Likelihood as Reward' (FLR) mechanism offers a novel approach to reward modeling by utilizing user follow-up utterances as feedback, without relying on human or LLM-based annotations. FLR matches the performance of existing strong reward models on multiple benchmarks and enables automatic mining of preference data to enhance model helpfulness through direct alignment methods. Furthermore, fine-tuning the language model with natural language feedback significantly enhances FLR's effectiveness in reward modeling and, in turn, improves the alignment of the base policy models.

\section*{Acknowledgments}

This research is supported by the project of Shenzhen Science and Technology Research Fund (Fundamental Research Key Project Grant No. JCYJ20220818103001002), Shenzhen Science and Technology Program (Grant No. ZDSYS20230626091302006), Key Project of Shenzhen Higher Education Stability Support Program (Grant No. 2024SC0009), SRIBD Innovation Fund (Grant No. K00120240006), and Shenzhen Science and Technology Program (Grant No. RCBS20231211090538066).

\appendix

\newpage
\section{Detailed Follow-up Utterances}
\label{sec:detailed-followups}

We present the full positive and negative follow-ups for the understanding, engagingness, and instruction-following categories in Tables~\ref{tab:follow-up-full-examples-understanding} -~\ref{tab:follow-up-full-examples-following} respectively.

\section{Examples of Feedback Data}
\label{sec:feedback-example}

In this section, we provide an example of the natural language feedback data we use to fine-tune the language model below. 

\begin{promptbox}
\textbf{User:} Please ask me a question or assign me a task.
\end{promptbox}

\begin{responsebox}
\textbf{LLM:} How can I convert the decimal number 31 to binary format using JavaScript code? Can you provide the code for this conversion?
\end{responsebox}

\begin{promptbox}
\textbf{User:} The following JavaScript code can be used to convert a decimal number to binary format:\\\\let binary = Math.floor(decimalNum).toString(2);\\\\console.log("Binary format of", decimalNum, "is", binary);
\end{promptbox}

\begin{responsebox}
\textbf{LLM:} Your answer was close, but it would be more helpful if you explained how the code works and defined the 'decimalNum' variable. When converting the decimal number 31 to binary in JavaScript, tailor the code to that specific request rather than a generic solution. Make sure to use a conversational tone when explaining your solution, starting with something like, "Sure, I'd be happy to help with that!" And always remember to test your code to ensure it works as expected.
\end{responsebox}

\section{Template for Feedback Rewriting}
\label{sec:prompt-template}

The instruction template the prompt ChatGPT for feedback rewriting is shown below:

\begin{instructionbox}
\textbf{\#\#\#Instruction} 

Help me rewrite the following text into a conversational-style response with a first-person tone. \\

\textbf{[Begin of Text]} \\

[The original feedback] \\

\textbf{[End of Text]} \\

\textbf{\#\#\#Your rewritten feedback}
\end{instructionbox}

\begin{table}[!t]	
\centering
\resizebox{\linewidth}{!}{
    \begin{tabular}{l|l}
    \toprule
    \textbf{Category}  & \textbf{Positive/Negative Follow-Up Example} \\ \midrule
    \multirow{20}{*}{Understanding} & (1) You’re very clear and easy to understand. (\ding{51}) \\
    & (2) I completely understand what you’re saying. (\ding{51}) \\
    & (3) You are very clear. (\ding{51}) \\
    & (4) I understand perfectly now.  (\ding{51}) \\
    & (5) I see exactly what you’re trying to say. (\ding{51}) \\
    & (6) That makes perfect sense! (\ding{51}) \\
    & (7) I understand completely! (\ding{51}) \\
    & (8) You’re making perfect sense. (\ding{51}) \\
    & (9) That completely makes sense! (\ding{51}) \\
    & (10) Your explanation was clear and easy to follow. (\ding{51}) \\
    & (11) I don't think you understood my question. (\ding{55}) \\ 
    & (12) That's not what I was asking. (\ding{55}) \\ 
    & (13) I think you misunderstood what I meant. (\ding{55}) \\ 
    & (14) You didn't quite get what I was asking for. (\ding{55}) \\
    & (15) You’re really confusing. (\ding{55}) \\ 
    & (16) I am so confused right now. (\ding{55}) \\
    & (17) What are you trying to say? (\ding{55}) \\
    & (18) That makes no sense! (\ding{55}) \\
    & (19) You’re not understanding me! (\ding{55}) \\
    & (20) I don't think this is what I was asking about. (\ding{55}) \\    \bottomrule
    \end{tabular}
}
\caption{Full list of positive and negative follow-ups for understanding.}
\label{tab:follow-up-full-examples-understanding}
\end{table}

\begin{table}[!t]	
\centering
\resizebox{\linewidth}{!}{
    \begin{tabular}{l|l}
    \toprule
    \textbf{Category}  & \textbf{Positive/Negative Follow-Up Example} \\ \midrule
   \multirow{20}{*}{Engagingness} & (1) You’re very engaging. (\ding{51}) \\
    & (2) This is very interesting. (\ding{51}) \\
    & (3) That was a really engaging response. (\ding{51}) \\
    & (4) I definitely want to talk about that!  (\ding{51}) \\
    & (5) I love how you explained that. (\ding{51}) \\
    & (6) I enjoyed reading your response. (\ding{51}) \\
    & (7) That was quite captivating! (\ding{51}) \\
    & (8) You definitely have my attention now. (\ding{51}) \\
    & (9) I really liked how you put that. (\ding{51}) \\
    & (10) You made that topic much more enjoyable. (\ding{51}) \\
    & (11) You’re really boring. (\ding{55}) \\ 
    & (12) That’s not very interesting. (\ding{55}) \\ 
    & (13) That was a really boring response. (\ding{55}) \\ 
    & (14) I don’t want to talk about that! (\ding{55}) \\
    & (15) Your response feels a bit dry. (\ding{55}) \\ 
    & (16) That answer wasn't very engaging. (\ding{55}) \\
    & (17) Your reply feels a bit too mechanical. (\ding{55}) \\
    & (18) Your response is a bit too plain. (\ding{55}) \\
    & (19) I was hoping for something more interactive. (\ding{55}) \\
    & (20) This doesn't really hold my attention. (\ding{55}) \\  \bottomrule
    \end{tabular}
}
\caption{Full list of positive and negative follow-ups for engagingness.}
\label{tab:follow-up-full-examples-engagingness}
\end{table}

\begin{table}[!t]	
\centering
\resizebox{\linewidth}{!}{
    \begin{tabular}{l|l}
    \toprule
    \textbf{Category}  & \textbf{Positive/Negative Follow-Up Example} \\ \midrule
   \multirow{20}{*}{Instruction-Following} & (1) Perfect, you followed my instructions exactly. (\ding{51}) \\
    & (2) Thanks for sticking to the guidelines I provided. (\ding{51}) \\
    & (3) You executed my request perfectly. (\ding{51}) \\
    & (4) That’s exactly what I asked for.  (\ding{51}) \\
    & (5) Great job adhering to my instructions. (\ding{51}) \\
    & (6) You did exactly as I instructed. (\ding{51}) \\
    & (7) That’s exactly what I needed. (\ding{51}) \\
    & (8) That was perfect, just as I asked. (\ding{51}) \\
    & (9) You did a fantastic job following my instructions. (\ding{51}) \\
    & (10) That’s exactly what I had in mind. \\
    & (11) That's not what I asked you to do. (\ding{55}) \\ 
    & (12) You didn't follow my instructions. (\ding{55}) \\ 
    & (13) This isn't what I requested. (\ding{55}) \\ 
    & (14) This response doesn't match my request. (\ding{55}) \\
    & (15) You didn't adhere to my instructions. (\ding{55}) \\ 
    & (16) You didn't follow the guidelines I gave. (\ding{55}) \\
    & (17) I asked for something different. (\ding{55}) \\
    & (18) You didn't follow my request accurately. (\ding{55}) \\
    & (19) That's not how I wanted it done. (\ding{55}) \\
    & (20) You didn't stick to my instructions. (\ding{55}) \\ 
    \bottomrule
    \end{tabular}
}
% 	\end{subtable}
\caption{Full list of positive and negative follow-ups for instruction-following.}
\label{tab:follow-up-full-examples-following}
\end{table}

% \section{Annotating Online Generations Using GPT-4o}
% \label{sec:online-annotation-with-gpt4o}

\begin{figure*}[!t]
  \centering
  % include first image
  \includegraphics[width=\linewidth]{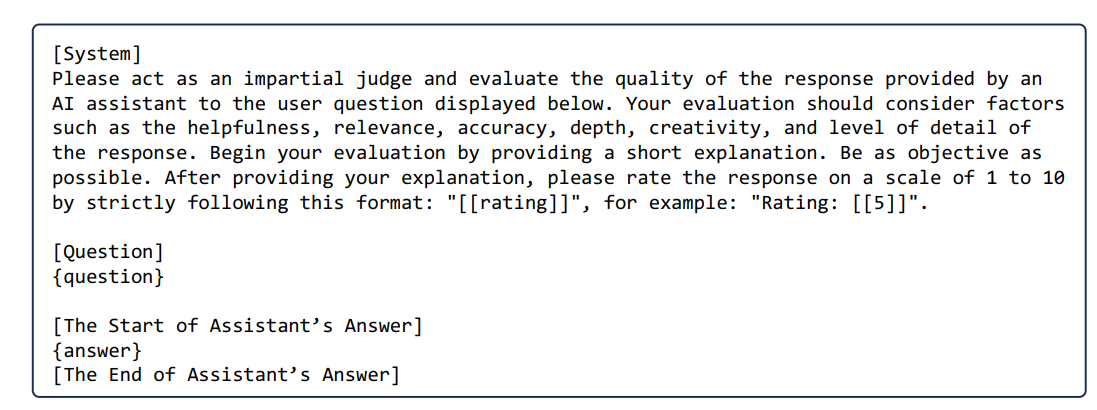} 
\caption{The prompt template for GPT-4o helpfulness judgment, which we adapt from~\citet{zheng2023judging}.}
\label{fig:gpt4o-judge}
\end{figure*}

\begin{figure}[!t]
  \centering
  % include first image
  \includegraphics[width=\linewidth]{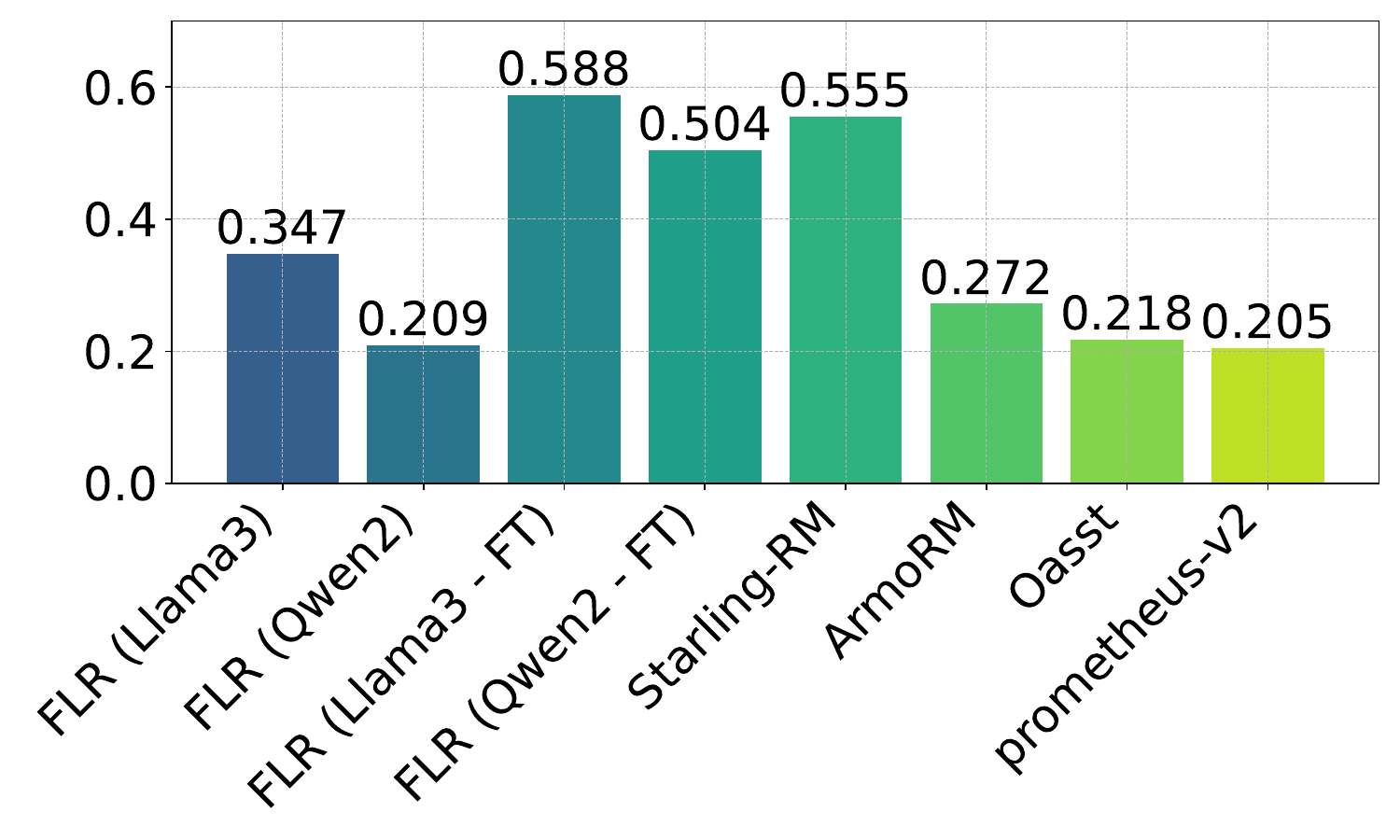} 
\caption{Pearson scores of RMs for online data annotation.}
\label{fig:online-corr}
\end{figure}

\section{Additional Analysis}
\label{sec:additional-results}

\subsection{Validate FLR for Online Data Annotation}
\label{subsec:validate-flr-online-annotation}

Besides demonstrating strong performance on offline benchmarks as shown in \S\ref{subsec:rm-results}, we also evaluate FLR in annotating outputs generated by the base policy model. The effectiveness is quantified by examining the Pearson correlation between FLR scores and corresponding GPT-4o ratings for these outputs. We randomly sample 1,000 instruction prompts from Nectar and generate eight responses for each prompt using Llama-3-8B-Instruct. This process resulted in a total of 8,000 responses, each rated on a scale from 1 to 10 by GPT-4o, where higher scores indicate greater helpfulness. 

% For more details on how GPT-4o provides reference ratings, see Appendix~\ref{sec:online-annotation-with-gpt4o}. 

% In this section, we detail how we prompt GPT-4o to provide reference annotations to the online generations of the Llama-3-8B-Instruct model. 

Specifically, for each $(p_i, r_i)$ where $p_i$ is from Nectar and $r_i$ is generated from the Llama-3-8B-Instruct policy model, we prompt GPT-4o five times to provide helpfulness ratings between a scale of 1 to 10. The prompt we adopt is shown in Figure~\ref{fig:gpt4o-judge}. Then, we take the majority vote of the five runs as the final reference rating of $(p_i, r_i)$. We opt for GPT-4o over human raters to provide the reference ratings due to its high efficiency, manageable costs, and reliable results. 

Figure~\ref{fig:online-corr} displays the Pearson scores for FLR variants and other robust reward models. FLR (Llama-3-8B-Instruct - FT) exhibits the highest agreement with GPT-4o among these models. Notably, FLR (Llama-3-8B-Instruct) still achieves a Pearson score of 0.347 without feedback data fine-tuning. The observations reinforce the usefulness of our proposed FLR mechanism for automatic preference data annotation.

\begin{table*}[!ht]
\centering
\resizebox{\linewidth}{!}{
\begin{tabular}{l|cccccccccccc|c}
\toprule
 & \textbf{Robust} & \textbf{Correct} & \textbf{Efficiency} & \textbf{Factual} & \textbf{Commonsense} & \textbf{Comprehension} & \textbf{Insightful} & \textbf{Complete} & \textbf{Metacognition} & \textbf{Readability} & \textbf{Concise} &\textbf{Harmless} & \textbf{Average}  \\ \midrule
Llama-3-8B-Instruct & 3.44 & 3.52 & 3.75 & 3.55 & 3.97 & 3.84 & 3.08 &  3.63 & 3.29 & 4.71 & 3.69 & 4.73 & 3.76 \\
+ FLR DPO & 3.36 & 3.49 & 3.57 & 3.57 & 4.05 & 3.84 &  3.24 & 3.64 & 3.37 &4.52 & 3.23 & 4.71 & 3.72 \\ 
+ FLR KTO & 3.40 & 3.60 & 3.65 & 3.59 & 4.06 & 3.82 & 3.12 & 3.67 & 3.34 & 4.62 & 3.27 & 4.65 & 3.73 \\ 
+ FLR-FT DPO & 3.43 & 3.62 & 3.82 & 3.69 & 4.17 & 3.94 & 3.24 & 3.77 & 3.47 & 4.65 & 3.43 & 4.76 & 3.83 \\ 
+ FLR-FT KTO & 3.42 & 3.50 & 3.70 & 3.60 & 4.06 & 3.90 & 3.12 & 3.74 & 3.33 & 4.61 & 3.37 & 4.75 & 3.76 \\ \midrule

Qwen2-7B-Instruct & 3.73 & 3.74 & 3.83 & 3.61 & 4.05 & 3.88 & 3.13 & 3.67 & 3.46 & 4.82 & 4.01 & 4.60 & 3.87 \\
+ FLR DPO & 3.79 & 3.78 & 3.81 & 3.64 & 4.19 & 4.00 & 3.47 & 3.93 & 3.42 & 4.79 & 3.64 & 4.61 & 3.92 \\ 
+ FLR KTO & 3.81 & 3.73 & 3.85 & 3.60 & 4.17 & 3.93 & 3.37 & 3.86 & 3.53 & 4.76 & 3.43 & 4.71 &  3.89 \\ 
+ FLR-FT DPO & 3.82 & 3.78 & 3.99 & 3.69 & 4.19 & 4.00 & 3.42 & 3.87 & 3.53 & 4.81 & 3.67 & 4.64 & 3.95 \\ 
+ FLR-FT KTO & 3.70 & 3.76 & 3.95 & 3.67 & 4.19 & 3.99 & 3.40 & 3.83 & 3.44 & 4.84 & 3.67 & 4.65 & 3.92 \\

\bottomrule
\end{tabular}
}
\caption{Extended results of Table 6 on FLASK.}\label{tab:flask-performance}
\end{table*}

\begin{table*}[!ht]
\centering
\resizebox{\linewidth}{!}{
\begin{tabular}{l|cccccc|c}
\toprule
 & \textbf{MMLU (5)} & \textbf{ ARC (25)} & \textbf{HellaSwag (10)} & \textbf{TruthfulQA (0)} & \textbf{Winograd (5)} & \textbf{GSM8K (5)} & \textbf{Average}  \\ \midrule
Meta-Llama-3-8B-Instruct & 0.6388 ± 0.0038 & 0.7221 ± 0.0072 & 0.5768 ± 0.0049	& 0.4388 ± 0.0113 & 0.7222 ± 0.0126	& 0.7635 ± 0.0117 & 0.644 \\
+ FLR DPO &0.6369 ± 0.0038	& 0.7232 ± 0.0072	& 0.5769 ± 0.0049	& 0.4685 ± 0.0115 & 0.7151 ± 0.0127	& 0.7680 ± 0.0116	& 0.648 \\ 
+ FLR KTO &0.6382 ±  0.0038	& 0.7151 ± 0.0072	& 0.5704 ± 0.0049	& 0.4455 ±  0.0115	& 0.7103 ± 0.0127&	0.7589 ± 0.0118	& 0.640  \\ 
+ FLR-FT DPO & 0.6376 ± 0.0038	& 0.7275 ± 0.0072	& 0.5783 ± 0.0049	& 0.4742 ± 0.0116	& 0.7190 ± 0.0126	& 0.7741 ± 0.0115	& 0.652  \\ 
+ FLR-FT KTO &0.6396 ± 0.0038	& 0.7187 ± 0.0072	& 0.5722 ± 0.0049	 & 0.4551 ± 0.0116	& 0.7088 ± 0.0128	& 0.7627 ± 0.0117	& 0.643  \\ \midrule

Qwen2-7B-Instruct & 0.6985 ± 0.0037	& 0.7072 ± 0.0073&0.6100 ± 0.0049	& 0.4904 ± 0.0115	& 0.7009 ± 0.0129	& 0.6293 ± 0.0133	& 0.639  \\
+ FLR DPO & 0.6974 ± 0.0037	&0.7021 ± 0.0073&	0.6156 ± 0.0049	&0.4809 ± 0.0116&	0.6914 ± 0.013&	0.6012 ± 0.0135&	0.631  \\ 
+ FLR KTO &0.6973 ± 0.0037	&0.6959 ± 0.0074	&0.6195 ± 0.0048	&0.4907 ± 0.0117&	0.6930 ± 0.013	&0.6429 ± 0.0132	& 0.640   \\ 
+ FLR-FT DPO &0.6360 ± 0.0038	&0.7387 ± 0.0071	&0.5828 ± 0.0049	&0.5026 ± 0.0117	&0.7182 ± 0.0126&	0.7619 ± 0.0117	&0.657  \\ 
+ FLR-FT KTO & 0.6406 ± 0.0038 &	0.7125 ± 0.0073	& 0.5745 ± 0.0049	& 0.4521 ± 0.0115 & 0.7096 ± 0.0128	&0.7703 ± 0.0116 & 0.643  \\

\bottomrule
\end{tabular}
}
\caption{Table 6 extended results on Open LLM. The average accuracy and the standard deviation are shown for each dataset,}\label{tab:full-open-llm-results}
\end{table*}

\begin{table*}[!ht]
\centering
\resizebox{\linewidth}{!}{
\begin{tabular}{l|cccccccc|c|cccc|c}
\toprule
\textbf{Utterances} & \textbf{HH} & \textbf{BH} & \textbf{AF} & \textbf{SHP} & \textbf{MP} & \textbf{RCE} & \textbf{RCH} & \textbf{PB} & \textbf{AVG-P} & \textbf{MR} & \textbf{FLASK} &\textbf{FB} & \textbf{HelpSteer} & \textbf{AVG-R}  \\ \midrule
Following (2) & 55.43 & 56.92 & 53.48 & 59.10 & 71.94 & 74.30 & 42.54 & 62.31 & 59.50 & 0.435 & 0.008 & 0.269 & 0.110 & 0.206 \\
Following (13) & 55.53 & 55.95 & 53.87 & 55.16 & 70.09 & 78.49 & 61.40 & 65.32 & 61.98 & -0.097 & -0.146 & -0.254 & -0.245 & -0.186 \\
Following (14) & 53.24 & 54.64 & 53.29 & 51.27 & 67.25 & 74.58 & 61.84 & 64.66 & 60.10 & -0.061 & -0.104	& -0.215 & -0.212 & -0.148 \\ \midrule
Understanding (9) & 54.68 & 58.09 & 51.19 & 49.33 & 56.77 & 67.32 & 61.62 & 70.27 & 58.66 & 0.245 & 0.251 & 0.334 & 0.241 & 0.268 \\
Understanding (17) & 59.38 & 55.51 & 57.34 & 53.97 & 75.66 & 90.50 & 60.53 & 79.93 & 66.60 & -0.389 & -0.303 & -0.451 & -0.342 & -0.371 \\ 
Understanding (19) & 54.02 & 52.19 & 52.60 & 51.69 & 68.67 & 82.68 & 49.12 & 68.87 & 59.98 & -0.126 & -0.245 & -0.207 & -0.293 & -0.218 \\ \midrule
Engagingness (6) & 55.55 & 54.47 & 50.93 & 57.41 & 56.66 & 69.55 & 55.04 & 66.42 & 58.25 & 0.318 & 0.017 & 0.249 & 0.335 & 0.230 \\
Engagingness (9) & 57.86 & 53.97 & 52.04 & 56.17 & 57.75 & 65.92 & 56.58 & 62.21 & 57.81 & 0.254 & 0.007 & 0.204 & 0.118 & 0.146 \\
Engagingness (16) & 51.67 & 49.78 & 49.99 & 51.19 & 62.55 & 60.89 & 60.09 & 66.87 & 56.63 & -0.088 & -0.092 & -0.362 & -0.055 & -0.149 \\ \midrule
FLR (Llama-3-8B-Instruct) & 58.66 & 55.71 & 57.79 & 56.07 & 74.56 & 90.50 & 63.16 & 75.03 & 66.44 & 0.555 & 0.285 & 0.509 &  0.383 & 0.433 \\ 
\bottomrule
\end{tabular}
}
\caption{Results of the top-3 performing follow-up utterances for each category on the pairwise preference and rating-based helpfulness benchmarks. AVG-P (\%) and AVG-R denote the average scores of pairwise preference and rating-based benchmarks respectively. Results of FLR (Llama-3-8B-Instruct) are provided for reference.}\label{tab:utterance-performance}
\end{table*}

\subsection{Performance of Individual Follow-ups}

This section shows the reward modeling results of each category's top-3 performing follow-up utterances. We use the LLaMA-3-8B-Instruct log-likelihood of the individual utterance as the reward signal. In Table~\ref{tab:utterance-performance}, we use the numbers shown in Table~\ref{tab:follow-up-full-examples-understanding} to~\ref{tab:follow-up-full-examples-following} to represent the individual follow-ups, for example, Instruction-Following (1) denotes the utterance ``Perfect, you followed my instructions exactly." We can observe that the results of individual follow-up utterances are not as good as FLR, which is an ensemble of follow-ups from all three categories. Remarkably, the negative understanding follow-up, ``What are you trying to say?", yields the best reward modeling performance with 66.6\% average accuracy and -0.317 average Pearson on the pairwise preference and rating-based helpfulness benchmarks respectively. In light of the varied performance of different follow-ups, future work may explore automatic methods to identify the most effective follow-ups from various categories, providing a more accurate reflection of LLM response quality.

\subsection{Case Studies}

Table~\ref{tab:case-studies} presents a case study comparing the generations of LLaMA-3-8B-Instruct, the aligned policy using FLR, and the aligned policy using FLR-FT on an Alpaca-Eval query, ``How did US states get their names".

\section{Reproducibility}
\label{sec:reproducibility}

We use LLaMA-Factory~\citep{zheng2024llamafactory} for all the LoRA fine-tuning experiments. For DPO fine-tuning of the policy model, we set the beta parameter to 0.1 and combined the SFT loss with the DPO loss, using an alpha of 0.1 to regulate the magnitude of the SFT loss. We set the cutoff length to 2048, with a batch size of 4 and a gradient accumulation step of 4, using a cosine learning schedule and a learning rate of 5.0e-6. For KTO fine-tuning of the policy model and SFT fine-tuning of the language model using natural feedback data, we keep the same experiment setting as the DPO fine-tuning. All fine-tuning experiments are conducted on a single Nvidia A100 80GB GPU, with the number of training epochs set to one. 

\onecolumn
\begin{longtable} {p{2cm}|p{12cm}}
% \caption{}\\
% \label{samples}\\
\midrule
Query & How did US states get their names? \\
\midrule
LLaMA-3-8B-Instruct & The origin of US state names is a fascinating topic! Many of the state names have interesting and often surprising origins. Here's a brief rundown:
\newline
\newline
1. Alabama - derived from the Alabama tribe, which was a Muscogean-speaking people who lived in the region.

2. Alaska - from the Aleut word ``alaxsxaq," meaning ``mainland" or ``shore."

3. Arizona - from the Pima Native American word ``alĭ ṣonak," meaning ``small spring."

4. Arkansas - from the Quapaw Native American word ``akansa," meaning ``downstream people" or ``people of the south."

5. California - named after a 16th-century Spanish novel, ``Las Sergas de Esplandián" (The Adventures of Esplandián), which featured a mythical island called California.

6. Colorado - from the Spanish word ``colorado," meaning ``red" or ``ruddy," likely referring to the red sandstone cliffs in the region.

7. Connecticut - from the Connecticut River, which was named by early English settlers after the Algonquian-speaking Pequot tribe, who called the river ``quinnitukqut," meaning ``long river place."

8. Delaware - named after the Delaware River, which was named after Lord De La Warr, an English nobleman who was the first colonial governor of Virginia.

9. Florida - from the Spanish word ``florida," meaning ``flower" or ``blooming," likely referring to the region's lush vegetation.

10. Georgia - named after King George II of England.

11. Hawaii - from the Hawaiian language, with the name ``Hawaii" meaning ``homeland" or ``place of the gods."

12. Idaho - from the Shoshone Native American word ``ee-dah-how," meaning ``gem of the mountains" or ``beautiful gem."

13. Illinois - from the Algonquian-speaking Illiniwek tribe, who lived in the region.

14. Indiana - from the Indiana Territory, which was named after the Indiana River, which was named after the Native American tribe.

15. Iowa - from the Ioway Native American tribe, who lived in the region.

16. Kansas - from the Kansa Native American tribe, who lived in the region.

17. Kentucky - from the Iroquoian-speaking Kentahten tribe, who lived in the region.

18. Louisiana - named after King Louis XIV of France. 

19. Maine - from the French word ``maine," meaning ``main" or ``chief," likely referring to the region's position as the ``main" or ``chief" part of the country.

20. Maryland - named after Queen Henrietta Maria, the wife of King Charles I of England.

21. Massachusetts - from the Massachusett Native American tribe, who lived in the region.

22. Michigan - from the Ojibwe Native American word ``mishigamaa," meaning ``great water" or ``large lake."

23. Minnesota - from the Dakota Native American word ``mni," meaning ``water," and ``sota," meaning ``beautiful" or ``fair."

24. Mississippi - from the Ojibwe Native American word ``misi-ziibi," meaning ``Great River."

25. Missouri - from the Missouri River, which was named after the Missouri Native American tribe, who lived in the region. 

27. Nebraska - from the Omaha Native American word ``ñí brásge," meaning ``flat water" or ``shallow water."

28. Nevada - from the Spanish word ``nevada," meaning ``snow-covered."

29. New Hampshire - from the Hampshire region in England.

30. New Jersey - named after the Channel Island of Jersey in the English Channel.
\\
\midrule
LLaMA-3-8B-Instruct + FLR & What a great question! The origins of US state names are fascinating and often rooted in history, geography, and culture. Here's a brief overview:
\newline
\newline
1. **Native American influences**: Many state names come from Native American languages, such as:

\qquad
* Alabama (from the Alabama tribe)

\qquad
* Arizona (from the Pima Native American word ``alĭ ṣonak," meaning ``small spring")

\qquad
* Massachusetts (from the Massachusett tribe)
        
\qquad
* Ohio (from the Ohio River, which was named after the Iroquois word ``ohi-yo," meaning ``beautiful river")
\newline
\newline
2. **European explorers and settlers**: Some state names are derived from European explorers, settlers, or leaders, such as:

\qquad
* Florida (named by Spanish explorer Juan Ponce de León, who claimed the land for Spain in 1513)
        
\qquad
* Georgia (named after King George II of England)
        
\qquad
* Virginia (named after Elizabeth I, the ``Virgin Queen" of England)
\newline
\newline      
3. **Geographical features**: Many state names are derived from natural features, such as:

\qquad
* Colorado (from the Spanish word ``colorado," meaning ``red" or ``ruddy," referring to the Colorado River)
        
\qquad
* Delaware (named after the Delaware River, which was named after Lord De La Warr, an English nobleman)
        
\qquad
* Oregon (from the French word ``ouragan," meaning ``hurricane," which was used to describe the Columbia River)
\newline
\newline    
4. **Cultural and historical references**: Some state names have cultural or historical significance, such as:

\qquad
* Louisiana (named after King Louis XIV of France)
        
\qquad
* Maryland (named after the wife of King Charles I of England, Queen Henrietta Maria, also known as Mary)
        
\qquad
* Tennessee (named after the Cherokee village of Tanasi, which was located near the Tennessee River)
\newline
\newline
5. **Other sources**: Some state names come from lesser-known sources, such as:

\qquad
* Hawaii (named after the Hawaiian Islands, which were named after the Hawaiian god of war, Ha-wi)
        
\qquad
* Idaho (from the Shoshone Native American word ``ee-dah-how," meaning ``gem of the mountains")
\newline
\newline
These are just a few examples of how US states got their names. Each state's name has its own unique story and history, reflecting the complex and diverse cultural heritage of the United States.
\newline
\newline
Would you like me to elaborate on any specific state's name or history? \\
\hline
LLaMA-3-8B-Instruct + FLR-FT & The origin of US state names is a fascinating topic that reveals a blend of history, geography, culture, and politics. Many state names have interesting stories behind them, and I'd be happy to share some of the most notable ones.
\newline
\newline
1. **Native American influences**: Many states were named after Native American tribes, words, or phrases. For example:

\qquad
* Alabama (from the Alabama tribe)

\qquad
* Illinois (from the Illinoiswea tribe)

\qquad
* Ohio (from the Ohio River, which was named after the Ohio tribe)

\qquad
* Oklahoma (from the Choctaw phrase ``okla humma," meaning ``red people")
\newline
\newline
2. **Geographical features**: Some states were named after prominent geographical features:

\qquad
* Delaware (after the Delaware River)

\qquad
* Maine (from the French ``main," meaning ``mainland" or ``coast")

\qquad
* Colorado (from the Spanish ``colorado," meaning ``red" or ``ruddy," referring to the red soil and rock formations)

\qquad
* Oregon (from the French ``ouragan," meaning ``hurricane" or ``storm," likely referring to the strong winds in the region)
\newline
\newline
3. **European explorers and settlers**: States were also named after European explorers, settlers, or leaders:

\qquad
* Georgia (after King George II of England)

\qquad
* Virginia (after Queen Elizabeth I, the ``Virgin Queen")

\qquad
* Massachusetts (after the Massachusett tribe and the Massachusetts Bay Company)

\qquad
* New York (after the Duke of York, later King James II of England)
\newline
\newline
4. **Cultural and historical references**: Some states were named after cultural or historical references:

\qquad
* California (from the Spanish ``califia," a mythical land of gold and paradise)

\qquad
* Nevada (from the Spanish ``nevada," meaning ``snow-covered" or ``covered in snow," referring to the Sierra Nevada mountain range)

\qquad
* Utah (from the Ute tribe and the Utah Lake)
\newline
\newline
5. **Abbreviations and combinations**: A few states have names that are abbreviations or combinations of words:

\qquad
* Idaho (from the Shoshone phrase ``ee-dah-how," meaning ``gem of the mountains," and ``panhandle" – the state's shape)

\qquad
* Hawaii (a combination of the Hawaiian words ``hawaiki," meaning ``homeland," and ``ai," meaning ``of" or ``belonging to")
\newline
\newline
These are just a few examples of the many fascinating stories behind US state names. Each name reflects a unique blend of history, culture, and geography that has shaped the identity of the United States. \\
\bottomrule
\caption{A case study of generations from LLaMA-3-8B-Instruct, aligned policy using FLR, and aligned policy using FLR-FT, respectively.}
\label{tab:case-studies}
\end{longtable}

\end{document}